\documentclass[preprint,12pt,authoryear]{elsarticle}

\usepackage{amssymb}
\usepackage{amsthm}

\usepackage[utf8]{inputenc} 
\usepackage[T1]{fontenc}    
\usepackage{hyperref}       
\usepackage{url}            
\usepackage{booktabs}       
\usepackage{amsfonts}       
\usepackage{nicefrac}       
\usepackage{xcolor}         

\usepackage{algorithm}
\usepackage{algorithmic}
\usepackage{amsmath}
\usepackage{mathrsfs}
\usepackage{amssymb}
\usepackage{caption}
\usepackage{subfig}
\usepackage{booktabs}       
\usepackage{graphicx}
\usepackage{multirow}	
\usepackage[normalem]{ulem}

\journal{arXiv}

\begin{document}
	
	\begin{frontmatter}
		
		
		
		\title{A Joint Time-frequency Domain Transformer for Multivariate Time Series Forecasting}
		
		\author[LabelTsinghua]{Yushu Chen}
		\author[LabelLSZ]{Shengzhuo Liu}
		\author[LabelTechorigin]{Jinzhe Yang}
		\author[LabelJH]{Hao Jing}
		\author[LabelTsinghua]{Wenlai Zhao}
		\author[LabelTsinghua]{Guangwen Yang}
		
		\affiliation[LabelTsinghua]{organization={Department of Computer Science and Technology, Tsinghua University},
			addressline={RM.3-126, FIT Building, Haidian District}, 
			city={Beijing},
			postcode={100084}, 
			country={China}}
		
		\affiliation[LabelLSZ]{organization={College of Computer Science and Mathematics, Fujian University of Technology},
			addressline={RM.213, Building C4}, 
			city={Fuzhou},
			state={Fujian},
			postcode={350118}, 
			country={China}} 
		
		\affiliation[LabelTechorigin]{organization={Techorigin},
			addressline={No.581, Jianzhu West Road, Binhu District}, 
			city={Wuxi},
			state={Jiangsu},
			postcode={214000}, 
			country={China}} 
		
		\affiliation[LabelJH]{organization={Earth System Modeling and Prediction Center},
			addressline={No.46, Zhongguancun South Street, Haidian District}, 
			city={Beijing},
			postcode={100081}, 
			country={China}} 	            
		
		
		\begin{abstract}
		In order to enhance the performance of Transformer models for long-term multivariate forecasting while minimizing computational demands, this paper introduces the Joint Time-Frequency Domain Transformer (JTFT). JTFT combines time and frequency domain representations to make predictions. The frequency domain representation efficiently extracts multi-scale dependencies while maintaining sparsity by utilizing a small number of learnable frequencies. Simultaneously, the time domain (TD) representation is derived from a fixed number of the most recent data points, strengthening the modeling of local relationships and mitigating the effects of non-stationarity. Importantly, the length of the representation remains independent of the input sequence length, enabling JTFT to achieve linear computational complexity. Furthermore, a low-rank attention layer is proposed to efficiently capture cross-dimensional dependencies, thus preventing performance degradation resulting from the entanglement of temporal and channel-wise modeling. Experimental results on six real-world datasets demonstrate that JTFT outperforms state-of-the-art baselines in predictive performance.
		\end{abstract}

		\fntext[LabelWork]{Yushu Chen, Wenlai Zhao, and Guangwen Yang also work at National Supercomputing Center in Wuxi, China. Guangwen Yang is the corresponding author.}
		\ead{ygw@tsinghua.edu.cn}
		
		
		
		
	\end{frontmatter}
	
	
	\section{Introduction}

Time series forecasting predicts the future based on historical data. It has broad applications including but not limited to climatology, energy, finance, trading, and logistics \citep{Petropoulos2022}. Following the great success of Transformers \citep{transformer} in NLP \citep{nlpsurvey}, CV \citep{cvsurvey}, and speech \citep{speechsurvey}, Transformers have been introduced in time series forecasting and achieves promising results \citep{tssurvey}. 

One of the primary drawbacks of Transformers is their quadratic complexity in both computation and memory, making them less suitable for long-term forecasting. To address this limitation, a plethora of Transformer-based models, e.g., LogTrans, Informer, AutoFormer, Performer, and PyraFormer \citep{logtrans, informer, autoformer, performer, pyraformer}, have been proposed to enhance predictive performance while maintaining low complexity.  Notably, \citet{fedformer} observed that most time series which are dense in the time domain (TD) tend to have a sparse representation in the frequency domain (FD). In response, they introduced FEDFormer, which leverages a randomly selected subset of frequency components to exploit FD sparsity. This approach has linear complexity and achieved state-of-the-art (SOTA) results in early 2022. However, the sparse representation using random frequencies alone may not fully capture the multi-scale characteristics of time series. As a result, seasonal-trend decomposition remains necessary in FEDFormer, despite its susceptibility to hyperparameters. Additionally, due to the periodic nature of the basis functions, FD representation is vulnerable to non-stationarity. Enhancing the sparse FD representation to capture multi-scale features has become crucial, especially after the emergence of a simple linear model known as DLinear, which outperformed existing Transformers in time-series forecasting \citep{dlinear}. PatchTST \citep{patchtst}, an improved Transformer with a patching approach, has recently outperformed DLinear in performance. While PatchTST's concept is valuable for improving FD Transformers, it still exhibits quadratic complexity and cannot utilize cross-channel correlations.

In this paper, we present a joint time-frequency domain Transformer (JTFT). It exploits the FD sparsity in time series using a small number of learnable frequencies and enhances the learning of local relations by incorporating a fixed number of the latest data points. A low-rank attention layer is also used to effectively extract cross-channel dependencies. The main contributions lie in four folds:

\begin{itemize}
	\item [$\bullet$] A customize discrete cosine transform (CDCT) is presented to compute customized FD components and enable learning of frequencies. Based on CDCT, a FD representation with a few learnable frequencies is developed, which is effective to capture multi-scale structures of time series. 
	
	\item [$\bullet$] Time series are encoded based on both the sparse FD and latest TD representation, that enables the Transformer to extract both long-term and local dependencies effectively with linear complexity.
	
	\item [$\bullet$] A low-rank attention layer is introduced to learn cross-channel interaction, that further improves the predictive performance by mitigating the entanglement and redundancy in the capture of temporal and channel dependencies.
	
	\item [$\bullet$] Extensive experimental results on 6 real-world benchmark datasets covering multiple fields (medical care, energy, trade, transportation, and weather) show that our model improves the predictive performance of SOTA methods. Specifically, JTFT ranks as the top-performing model in 54 out of 60 settings with varying prediction lengths and metrics, and ranks the second in the remaining ones in the experiments.
\end{itemize}

The source code of JTFT is available at \url{https://github.com/rationalspark/JTFT.git}.

\section{Related work}

Well-known traditional methods for time series forecasting include AR, ARIMA, VAR , GARCH\citep{TimeSeriesAnalysisBook}, kernel methods\citep{kernel-method}, ensembles\citep{ensemble-method}, Gaussian processes\citep{Gaussian-Process_0}, regime switching\citep{regim-switch}, and so on. Benefiting from the improvement in computing power and neural network structure, many deep learning models are proposed and often empirically surpassed traditional methods in predictive performance.

Various deep neural network architectures, such as recurrent neural networks (RNN), convolutional neural networks (CNN), graph neural networks (GNN), multi-layer perceptrons (MLP), and Transformers, have been widely applied in time series forecasting. RNN-type models \citep{LSTM, Da-RNN, DeepSS, DeepAR} were once the most popular choice, but they often encountered issues related to gradient vanishing or exploding \citep{Pascanu2012OnTD}, limiting their performance. CNN structures, on the other hand, excel at extracting local features from time series \citep{tcn, Borovykh2017, Sen2019, micn, timesnet}. However, they usually require many layers to effectively represent global relationships, mainly due to their limited receptive field size. GNNs are increasingly applied to enhance both temporal and dimensional pattern recognition in time series data \citep{Wu2020, Cao2020}. Recent advancements \citep{nhits, Li2023mts, tide, tsmixer} suggest that MLP-type structures remain competitive in forecasting.

Transformers are appealing in time series forecasting since their attention mechanism is effective to capture long-term temporal dependency. However, the complexity and memory requirement are quadratic in the sequence length, which hinders application in long-term modeling. Numerous methods are proposed to both improve the predictive performance and reduce the costs of Transformers. Typically, these methods have to exploit some form of sparsity. LogTrans \citep{logtrans} introduces attention layers with LogSparse design and achieves $O(L\log^2 (L))$ complexity, where $L$ is the input length. Informer \citep{informer} selects the top-k in attention matrix with a KL-divergence based method, and has $O(L\log L)$ complexity. Autoformer \citep{autoformer} obtains the same complexity using an Auto-Correlation mechanism. Several improved Transformers have achieved linear complexity. Among these methods, Linformer \citep{linformer} compresses the sequence by learnable linear projection; Luna \citep{luna} adopts a nested linear structure; Nystr\"{o}former \citep{nystroformer} applies Nystr\"{o}m approximation in the attention mechanism; Performer \citep{performer} leverages a positive orthogonal random features approach; Pyraformer \citep{pyraformer} introduces a pyramidal attention module to extract dependency at different resolutions; FEDformer \citep{fedformer} combines seasonal-trend decomposition and frequency enhanced structures to capture both the global and local dependencies. Numerous effective approaches have been proposed besides the efforts to reduce complexity. Non-stationary Transformers \citep{non-stationary_transformers} and Scaleformer \citep{scaleformer} provide generic frameworks to improve the accuracy of existing methods. Crossformer \citep{crossformer} develops two-stage attention layers to utilize cross-channel dependency. \citet{Zhou2022} show that a linear head may achieve better performance than the Transformer decoder.

However, recent work \citep{dlinear} shows a simple linear model (DLinear) outperforms the existing SOTA Transformer based methods, and often by a large margin. Additionally, \citet{Li2023mts} argue that attention is not necessary for capturing temporal dependencies. Indeed, PatchTST \citep{patchtst}, which uses continuous segments of series as the input of a vanilla Transformer and a channel-independence setting, beats DLinear on the standard multivariate forecasting benchmarks recently. The method still needs improvement in terms of computational efficiency and the extraction of cross-channel dependency.

Frequency domain (FD) methods are also widely explored in time series modeling. Many commonly used FD features are obtained through the Discrete Fourier Transform (DFT). For instance, \citet{autoformer} efficiently compute the auto-correlation function using the Fast Fourier Transform (FFT). \citet{Leethorp2022} significantly accelerate training with minor accuracy loss by replacing self-attention in a Transformer encoder with the Fourier Transform. \citet{sun2022} present a competitive FD neural network for univariate time series forecasting. Rather than using all FD components, exploiting FD sparsity is crucial for computational efficiency and noise suppression. FEDformer, for example, randomly selects a subset of FD components. Similarly, \citet{wang2023novel} develop a compressed sensing technique using random FD components. \citet{film} improves the Legendre memory model using low-frequency Fourier components and a low-rank approximation. \citet{deeptime} develops a concatenated Fourier features module for efficiently learning high-frequency patterns. Additionally, \citet{difformer} combines temporal and frequency streams for time series classification and regression, with FD components selected by a band-pass filter. In contrast to these approaches, JTFT introduces a custom discrete cosine transform (CDCT) to extract FD features. The Discrete Cosine Transform (DCT) is chosen for its superior energy compaction characteristics compared to DFT, allowing it to obtain a sparse FD representation of the input effectively. CDCT further extends DCT by enabling the learning of frequencies, improving the extraction of periodic dependencies that may not align with the uniform frequency grids of DCT bases.

Other methods, such as Short Time Fourier Transform and Discrete Wavelet Transform, are used to capture time-frequency features \citep[e.g., ][]{chaovalit2011discrete, stfnets, singh2006optimal, Wen2021, ding2022novel}. These methods excel in extracting information from non-stationary data due to their ability to adapt FD representations over time. However, time-frequency features include FD components for discrete time points, resulting in considerably larger sizes compared to FFT and DCT. Consequently, these methods often require more effective feature selection or compression techniques to counteract overfitting. In contrast, JTFT effectively addresses the challenges posed by non-stationarity through recent TD representations.

\section{Methods}

In this section, we will introduce (1) the overall structure of JTFT, (2) the sparse FD representation with learnable frequencies, (3) the low-rank attention layers to extract cross-channel dependencies, and (4) the complexity analysis of the proposed method.

\subsection{JTFT framework}

\textbf{Preliminary:} Multivariate time series forecasting predicts the future value of time series based on historical data. We denote the input series by $\mathbf{x}_{1:L}=\{\mathbf{x}_1,\cdots,\mathbf{x}_L\}$, where $L$ represents the look-back window (input length). This series has $D$ channels, and the $i$-th channel is denoted as $\mathbf{x}^i = \{x_{1}^i,\cdots,x_{L}^i\}$. The future values to be forecasted are represented as  $\mathbf{x}_{L+1:L+T}=\{\mathbf{x}_{L+1},\cdots,\mathbf{x}_{L+T}\}$, where $T$ is the target-window (prediction length). The model's task is to map $\mathbf{x}_{1:L}$ to $\mathbf{y}\in \mathbb{R}^{D \times T}$, which is an approximation of $\mathbf{x}_{L+1:L+T}$.

\begin{figure}
	\centering
	\includegraphics[width=0.8\linewidth]{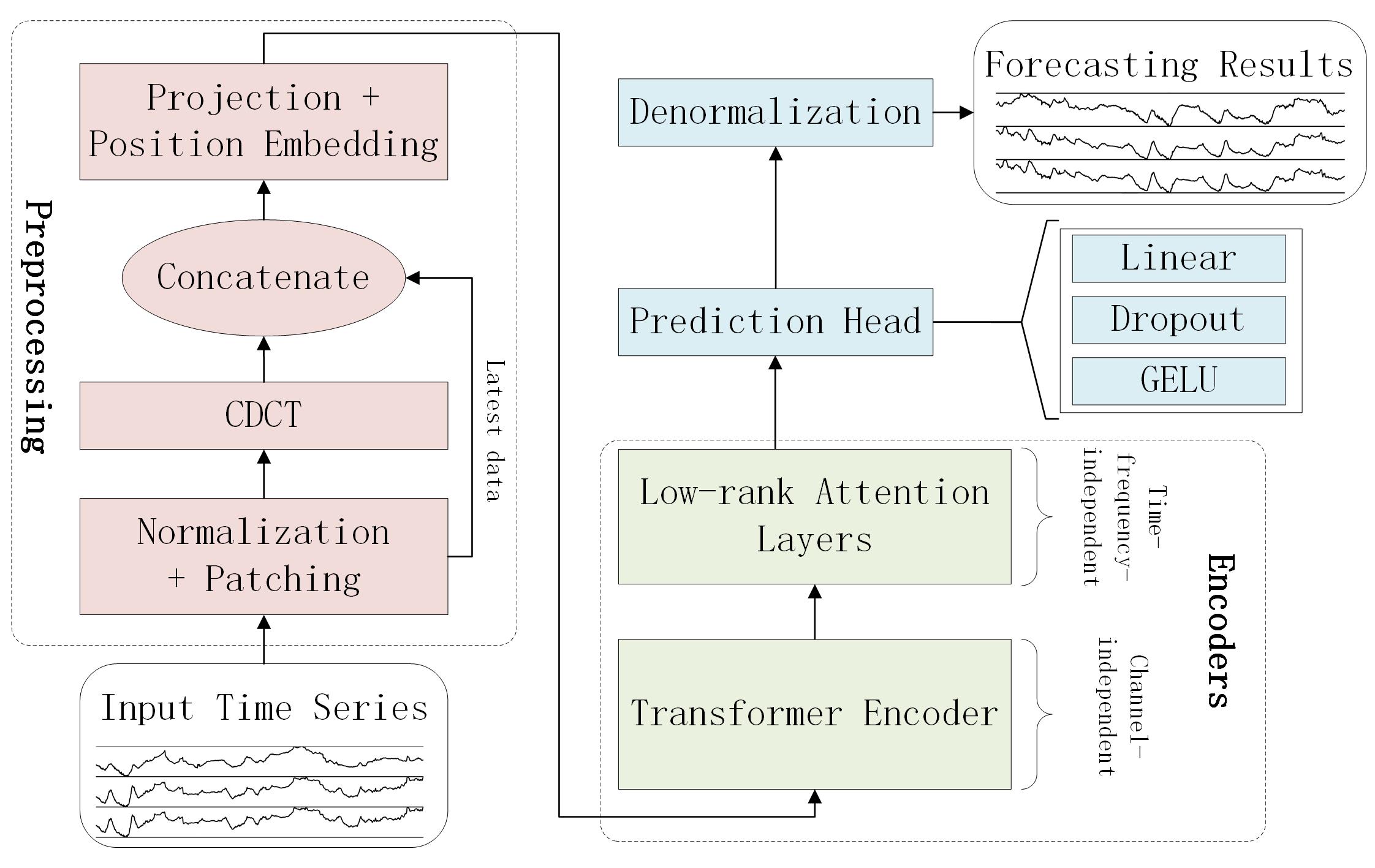}
	\caption{JTFT framework. The input time series undergo preprocessing to obtain a joint time-frequency embedding, which is then fed to the encoders for extracting time-frequency and cross-channel dependencies, respectively. The forecasting is generated using a prediction head.}
	\label{FIG_JTFT}
\end{figure}

\textbf{Overall structure:} The overall structure of JTFT is shown in Figure \ref{FIG_JTFT}. The model firstly transforms the input series to a joint time-frequency domain embedding, then maps it with a Transformer Encoder and a low-rank cross-channel attention mechanism to the latent representation, and finally generates the output using a prediction head followed by denormalization.

In the preprocessing stage, each channel of the input undergoes mean subtraction and standard deviation scaling. Afterward, a patching technique \citep{vit, beit, mae, patchtst} is applied, dividing the time series into either overlapped or non-overlapped segments (for detailed information, please refer to Appendix A). These continuous segments serve as the fundamental input units for the subsequent modeling stage.

\begin{figure}
	\centering
	\includegraphics[width=0.4\linewidth]{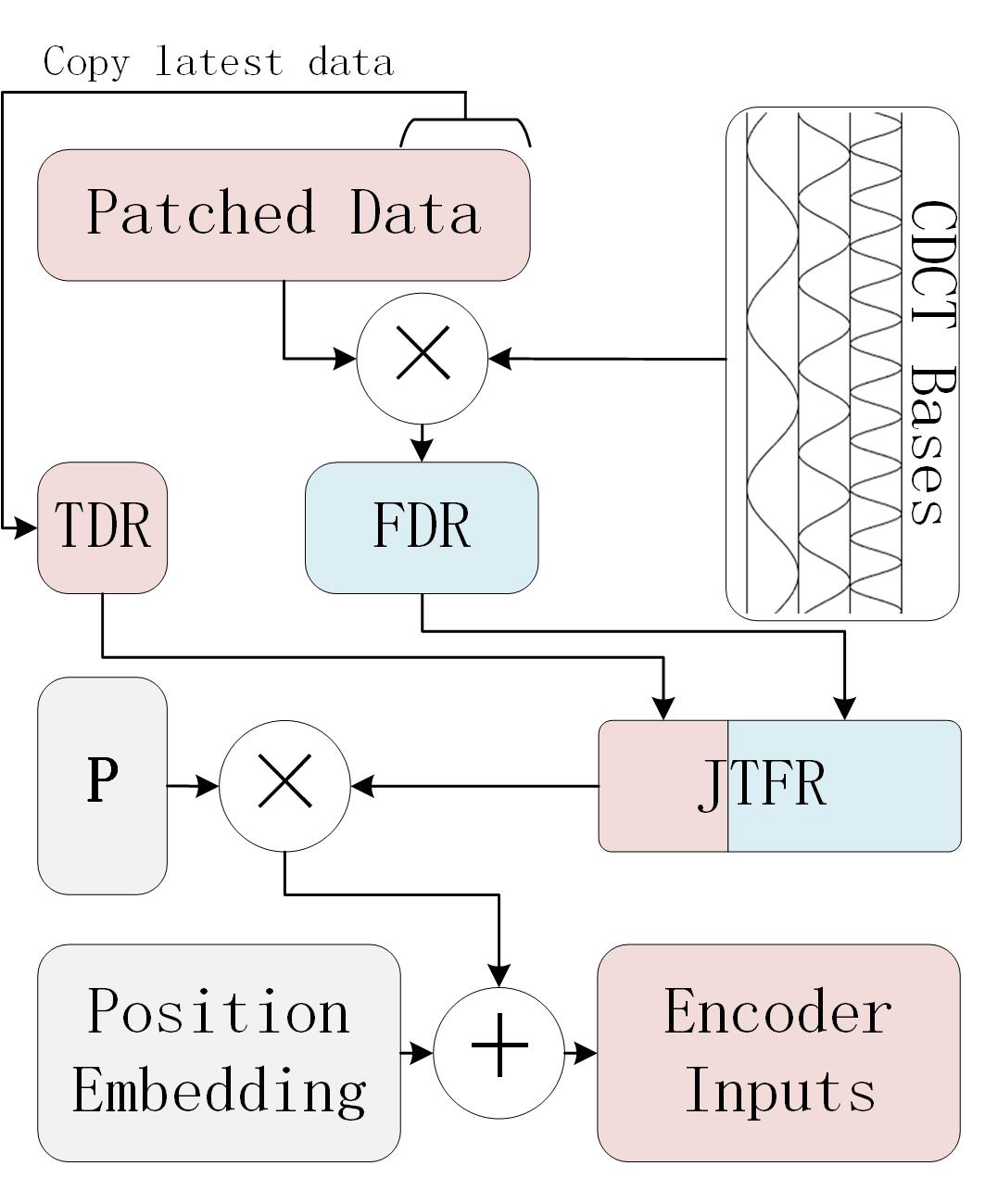}
	\caption{Time-Frequency Domain Data Embedding. Patched data is transformed into FD representation (FDR) by multiplying it with the CDCT bases. FDR is then concatenated with the TD representation (TDR) to form the JTFR. This JTFR is mapped to the model dimension using the projection matrix P and added with the position embedding to create the inputs for the encoders.}
	\label{FIG_EMBED}
\end{figure}

The preprocessed data then passes through a Custom Discrete Cosine Transform (CDCT) module to extract the frequency-domain (FD) components. These FD components are integrated with the most recent time-domain (TD) patches, creating a joint time-frequency domain representation (JTFR). Each channel's JTFR is subsequently projected into the latent space of the Transformer Encoder, forming an embedding of the input series, which is added by a learnable position embedding to account for the sequence's order. The progress is further illustrated in Figure \ref{FIG_EMBED}.

This fusion of TD and FD data plays a crucial role in mitigating the adverse effects of non-stationarity in time series data. Non-stationary data exhibits changes in statistical properties and joint distributions over time, making the time series less predictable \citep{non-stationary_transformers}. Additionally, the cyclical nature of the CDCT basis may not accurately represent these temporal changes. Therefore, the incorporation of the most recent TD data is essential for capturing up-to-date local relationships.

The Transformer encoder and low-rank attention layers in JTFT extract the time-frequency and cross-channel dependencies respectively. It is intuitively believed that cross-channel relationships are useful to improve prediction performance. However, \citet{patchtst} show that PatchTST with a channel-independent (CI) setting outperforms channel-mixing models, as it is less susceptible to overfitting and more robust to noise. Similarly, Crossformer \citep{crossformer} employs two-stage attention layers to learn cross-channel dependency, but its performance is inferior to PatchTST with CI in most of the experiments. Furthermore, \citet{Li2023mts} illustrate that entanglement and redundancy in capturing temporal and channel interaction affect forecasting performance. Motivated by these findings, JTFT leverages a two-stage approach to separately capture time-frequency and channel dependencies. The Transformer encoder in JTFT extracts time-frequency dependency using a CI setting. While cross-channel interaction is not modeled at this stage, the encoder is shared among all channels and trained with all available data, which helps to mitigate overfitting. The low-rank attention layers that extract cross-dimensional dependencies adopt a time-frequency-independent (TFI) setting, shared across all time-frequency points and trained with all available data, which also benefits generalization performance.

The prediction head consists of a GELU activation function \citep{gelu}, dropout layer \citep{dropout}, and linear projection layer. It maps the latent representation to output, which is then denormalized using the statistics saved in preprocessing. The model is trained using the Huber loss function, which offers greater resilience to outliers in the data compared to the Mean Squared Error (MSE) loss.

\subsection{The sparse FD representation with learnable frequencies}

Keeping all the frequency components may result in overfitting, because many high-frequency changes are caused by noises. It is also crucial to exploit FD sparsity in order to reduce the computation and memory complexity. Consequently, a critical problem for FD Transformers is how to select a subset of frequency components to represent the time series. Instead of keeping only the low frequency components, \citet{fedformer} show that the random selection used in FEDformer gives a better representation. They further show that the representing ability is close to the approximation by first $s$ largest single value decomposition, when the number of components is on the order of $s^2$. However, when a more precise frequency domain representation is expected, the $O(s^2)$ number of random components will be large. In order to reprensent the time series more precisely by less components, we introduce the FD representation with learnable frequecies.

A customize discrete cosine transform (CDCT) is developed to compute customized FD components, which enables learning of frequencies. The CDCT is a gerneralization of the discrete cosine transform (DCT), which expresses a sequence in terms of the sum of real-valued cosine functions oscillating at different frequencies. Compared with discrete Fourier transform (DFT) which uses complex exponential functions, DCT roughly halves the length to represent a real sequence, and has a strong energy compaction property. Consequently, it is more preferred in compression and thus suitable to represent series with a small number of FD components. A commonly used form of DCT is
\begin{equation}
	\begin{aligned}
		&\mathbf{\tilde{z}}=\text{DCT} (\mathbf{z})=\mathbf{\tilde{T}} \mathbf{z} \\
		&\mathbf{\tilde{T}}_{k, n}=\left\{\begin{aligned}
			&1/\sqrt{N}, \quad k=0, n\in\{0, \cdots, N-1\} \\
			&\sqrt{\frac{2}{N}} \cos \left(\frac{\pi}{N}\left(n+\frac{1}{2}\right) k\right), \quad k \in\{1, \cdots, N-1\}, n\in\{0, \cdots, N-1\},
		\end{aligned}\right. \\
	\end{aligned}
	\label{DCT}
\end{equation}
where $\mathbf{z}$ is a sequence with length $N$. It may differ from the input time sequence due to certain preprocessing steps. 

The matrix $\mathbf{\tilde{T}}$ is orthogonal, so the inverse transform is
\begin{equation}
	\mathbf{z}=\text{IDCT} (\mathbf{\tilde{z}})=\mathbf{\tilde{T}}^{-1} \mathbf{\tilde{z}}=\mathbf{\tilde{T}}^T \mathbf{\tilde{z}}.
	\label{IDCT}
\end{equation}
Some insignificant FD components in DCT can be ignored to exploit sparsity.

In definition (\ref{DCT}), the TD series $\mathbf{z}$ is transformed to FD by multiplying a orthogonal basis of cosine functions, the frequencies of which lie on uniform grid points indexed by $k$. However, these specified frequencies may not be adequate to express some real-world phenomenon. For example, when the sampling rate is 1, multiple FD components are required to express a simple function $\cos(1.1\pi t/N)$, but there is only one FD component in fact.

In order to further improve the representing ability with a small fixed number of FD components, we propose the CDCT by generalizing the DCT with customized frequencies as
\begin{equation}
	\begin{aligned}
		&\mathbf{\hat{z}}=\text{CDCT} (\mathbf{z})=\mathbf{\hat{T}} \mathbf{z} \\
		&\mathbf{\hat{T}}_{k, n}=\left\{\begin{aligned}
			&1/\sqrt{N}, \quad k=0, n\in\{0, \cdots, N-1\} \\
			&\sqrt{\frac{2}{N}} \cos \left(\left(n+\frac{1}{2}\right) \pi \psi_k\right), \psi_k \in (0,1),
		\end{aligned}\right. \\
	\end{aligned}
	\label{CDCT}
\end{equation}
where $\Psi=\{0, \psi_1, \cdots, \psi_{n_f-1}\}$ is a group of customized frequency coefficients, $k \in\{1, \cdots, n_f-1\}$, and $n\in\{0, \cdots, N-1\}$. We set $\psi_0=0$ to retain the mean. The CDCT maps a time series of length $N$ to $n_f$ FD components, that is applicatable to obtain a compact data representation in time series forecasting. If $n_f \ll N$ is treated as a constant, the complexity of the CDCT is $O(N)$. However, unlike the DCT, the CDCT does not have an intuitive inverse transform due to the fact that its basis is not orthogonal in general. Benefitting from modern deep learning frameworks, it is convenient to learn a projection matrix to recover the TD series from the sparse FD representation obtained by CDCT.

The CDCT allows the learning of frequencies by setting $\psi_{1:n_f-1}$ as learnable parameters. This flexibility enables the adjustment of frequencies within CDCT to better approximate the most significant frequencies in the data, which may not align with the uniform grid points of the DCT. In the implementation, $\psi_{1:n_f-1}$ are initialized based on the top frequencies obtained by applying DCT to a randomly sampled subset of the dataset. Although initializing CDCT with DCT frequencies has a time complexity of $O(L\log L)$, this initialization stage is relatively quick, particularly when compared to the training times. Since time series datasets are typically not as large as those in CV and NLP, even performing DCT on the whole datesets without sampling is acceptable in many real-world scenarios. Therefore, to be less rigorous, this short initialization stage is omitted when analyzing the overall complexity.

\begin{figure}
	\centering
	\includegraphics[width=\linewidth]{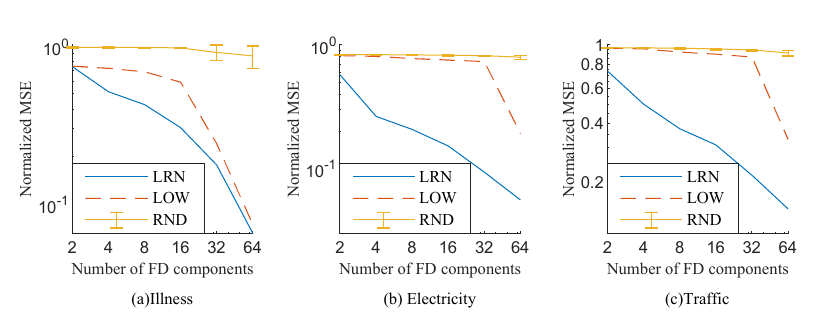}
	\caption{Comparison of the errors in reconstructing the input TD series from the FD representation of learnable (LRN), low (LOW), and random (RND) frequencies. The datasets are divided into continuous multivariate subsequences, each with a length of 512, and the MSE is normalized by the sum-of-squares of the data. The RND frequencies are executed 5 times, and the error bar represents the standard deviation.}
	\label{FIG_CDCT}
\end{figure}

We compare the representation ability of learnable (LRN), low (LOW), and random (RND) frequencies by examining the errors in reconstructing the input TD series. For random and low frequencies, the TD series is reconstructed using IDCT, while a learned linear projection is used for learnable frequencies. From the results shown in Figure \ref{FIG_CDCT}, the learnable frequencies representation retains significantly more information from the input compared to random and low frequencies when the number of components is the same. It's worth noting that lossless data representations of the inputs are unnecessary because insignificant frequency components often have higher noise levels and may be detrimental for prediction. Although the representational ability of low frequencies may be close to that of learnable frequencies when the number of components is large, the latter are more effective at obtaining a sparse FD representation with fewer components that retains the important information.

\subsection{The low-rank attention layers to extract cross-channel dependencies}

There are two intuitive ways to capture cross-channel dependencies. The first approach involves embedding data points from all channels at the same time step into a unified feature vector, while the second approach employs a Transformer along the channel dimension. \citet{crossformer} reduces the computational and memory costs in the channel-wise Transformer by replacing the self-attention with two smaller-scaled attentions. However, empirical results \citep{patchtst, Zhou2022} indicate that these approaches generate larger errors when compared to channel-independence (CI) PatchTST.

To address the need for capturing cross-channel dependencies with less redundancy, we introduce low-rank attention (LRA) layers. LRA is a computationally efficient approach to integrate cross-channel information into CI modeling. It conducts a lightweight attention across the channel dimension, generating low-rank corrections to the outputs of CI Transformer encoders.

The goal of LRA is to enhance accuracy beyond the already strong baseline provided by the CI model, which captures high-accuracy temporal dependencies. The process is prone to overfitting. LRA mitigates overfitting by confining the updates to a low-rank space through two approaches. First, it maps cross-channel sequences to low-dimensional representations, followed by a linear projection back to the original space to generate updates. Second, it employs a time-frequency independence (TFI) setting that shares corrections across the time-frequency dimension. Additionally, it reduces the number of parameters by simplifying the attention mechanism and moving the positional embedding from the input space to the low-dimensional space of representations.

\begin{figure}
	\centering
	\includegraphics[width=0.4\linewidth]{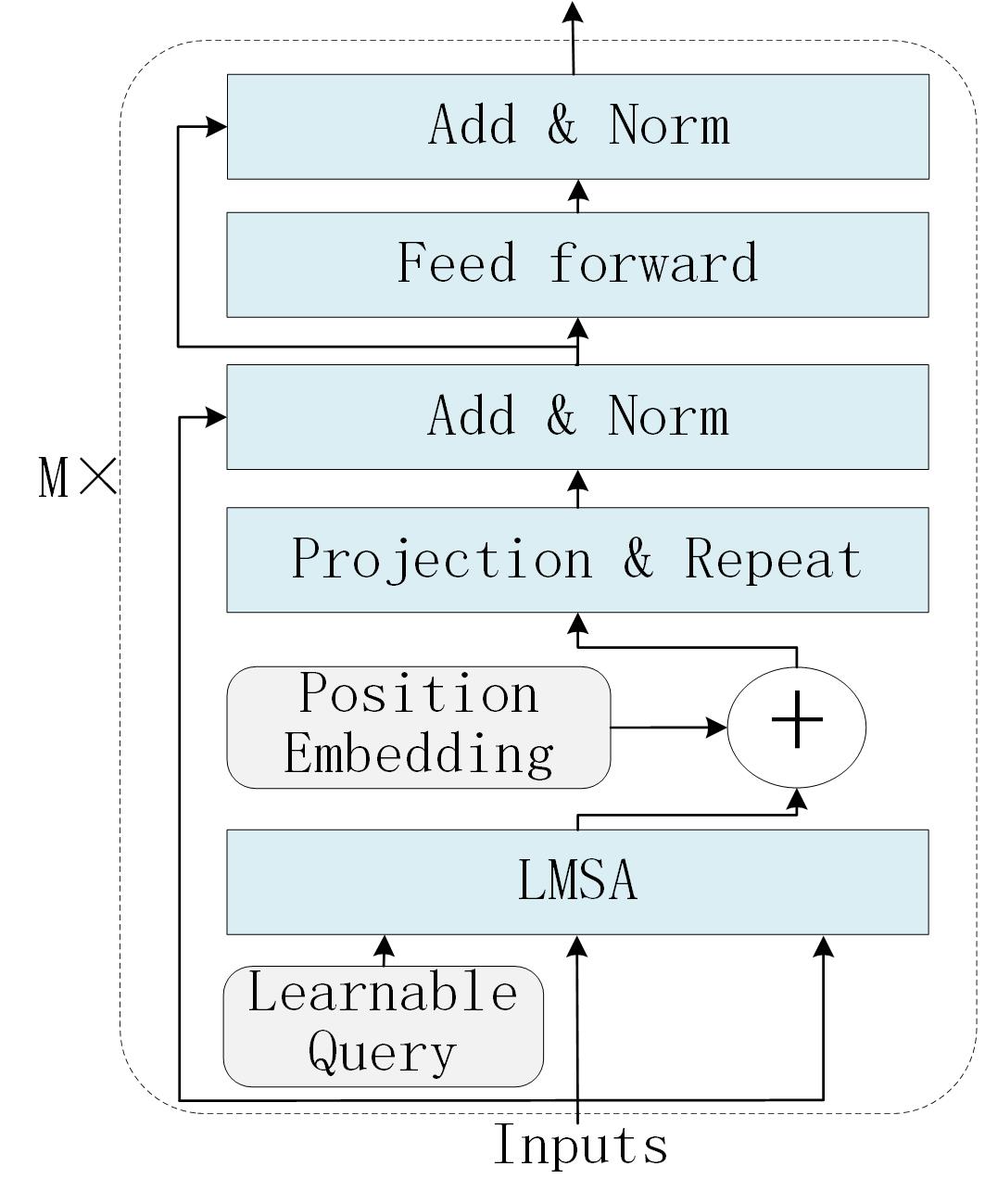}
	\caption{LRA Layers. It conducts attention across the channel dimension. Inputs and a learnable query are fed into LMSA, with the outputs being added with a compact learnable position embedding. The outcomes are projected across the channel dimension and replicated along the time-frequency dimension to match the input shape. The shortcut connection, LayerNorm, and MLP are applied similarly to the Transformer. M is the number of layers.}
	\label{FIG_LRA}
\end{figure}

Figure \ref{FIG_LRA} illustrates the structure of LRA. It replaces the resource-intensive multi-head self-attention (MSA) in a channel-wise Transformer encoder with a lightweight multi-head self-attention (LMSA). LMSA utilizes short learnable queries to aggregate messages from all channels into a low-rank space. Additionally, a compact learnable position embedding is added to the LMSA outputs to account for position within the low-rank space. The resulting outputs are mapped to match the number of channels through linear projection, distributing the aggregated messages among the channels. The results are duplicated in the time-frequency dimension, following the TFI setting. Subsequently, the shortcut connection, layer normalization (LayerNorm) \citep{layernorm}, and multi-layer feedforward network (denoted by MLP) are applied in the same manner as in the Transformer encoder.

The LMSA is a simplified version of the MSA with reduced computational requirements and parameters. In the LMSA, both the keys and values share the same linear projection. Because it is used alongside learnable queries, the linear projection for queries is also omitted. Denote the standard MSA by
\begin{equation}
	\text{Attn}(\mathbf{Q},\mathbf{K},\mathbf{V})=\text{Softmax}\left(\frac{\mathbf{Q}\mathbf{K}^T}{\sqrt{d_k}}\right)\mathbf{V},
	\label{ATTENTION}
\end{equation}
where $\mathbf{Q}\in \mathbb{R}^{l_q \times d_k}$,  $\mathbf{K}, \mathbf{V} \in \mathbb{R}^{l_{kv} \times d_k}$.
The LMSA is defined as
\begin{equation}
	\begin{aligned}
		&\text{LMSA}(\mathbf{\hat{Q}},\mathbf{\hat{K}},\mathbf{\hat{V}})=\text{Concat}(\text{head}_1, \cdots, \text{head}_h) \mathbf{W}_o \\
		&\text{head}_i=\text{Attn}\left(\mathbf{\hat{Q}}_{(i-1)d_k+1:i{d_k}}, \mathbf{\hat{K}}\mathbf{W}_{kv,i}, \mathbf{\hat{V}}\mathbf{W}_{kv,i}\right),
		\label{LMSA}
	\end{aligned}
\end{equation}
where the inputs $\mathbf{\hat{Q}}\in \mathbb{R}^{l_q \times h d_k}$, $\mathbf{\hat{K}}, \mathbf{\hat{V}} \in \mathbb{R}^{l_{kv} \times d_m}$, the learnable matrices $\mathbf{W}_o \in \mathbb{R}^{hd_{k} \times d_{m}}$, $\mathbf{W}_{kv,i} \in \mathbb{R}^{d_{m} \times d_{k}}$. $h$ is the number of heads, and $d_m$ is the model width. The batch-size dimension is ignored in the discussion for convenience.

The input of LRA is the CI representation obtained by the Transformer encoder. It is denoted as $\mathbf{Z}_{in} \in \mathbb{R}^{D \times (n_t+n_f) \times d_m}$, where $n_t$ and $n_f$ are the length of TD and FD representations. According to the TFI setting, $\mathbf{Z}_{in}$ is reshaped to $\mathbf{\hat{Z}} \in \mathbb{R}^{D (n_t+n_f)\times d_m}$. Then the LRA can be discribed as
\begin{equation}
	\begin{aligned}
		&\mathbf{B}=\text{LMSA}\left(\mathbf{R}, \mathbf{\hat{Z}}, \mathbf{\hat{Z}}\right)\\
		&\mathbf{\hat{B}}=\mathbf{B}+\mathbf{E}_{pos}\\
		&\mathbf{\bar{Z}}=\mathbf{W}_{e}\mathbf{\hat{B}}\\
		&\mathbf{\tilde{Z}}=\text{LayerNorm}\left(\mathbf{Z}_{in} \oplus \mathbf{\bar{Z}}\right)\\
		&\mathbf{Z}_{out}=\text{LayerNorm}\left(\mathbf{\tilde{Z}} + \text{MLP} \left(\mathbf{\tilde{Z}}\right)\right).
	\end{aligned}
	\label{LRA}
\end{equation}
In equation (\ref{LRA}), $\mathbf{R} \in \mathbb{R}^{d_r \times hd_k}$ is the learnable query, whose length $d_r \ll D$; $\mathbf{E}_{pos} \in \mathbb{R}^{d_r \times d_m}$ is the compact learnable position embedding; $\mathbf{W}_{e} \in \mathbb{R}^{D \times d_r}$ is the learnable projection matrix; $\oplus$ maps $\mathbf{\bar{Z}} \in \mathbb{R}^{D \times d_m}$ into $\mathbb{R}^{D \times (n_t+n_f) \times d_m}$ by repeating along the time-frequency dimension, and then adds it to $\mathbf{Z}_{in}$.

LRA is computationally efficient, benefiting from its low-rank nature.  While the channel-wise Transformer has a complexity of $O(D^2)$ with respect to $D$, which can lead to substantial computational costs, LRA reduces the complexity to $O(Dd_r)$. This complexity can be approximated as $O(D)$ when $d_r \ll D$ is viewed as a constant.

\subsection{Complexity analysis}

Instead of using the input series directly, most of the operations in JTFT are applied to the joint time-frequency domain representation. Its sequence length $\hat{L}$, which is the sum of the TD and FD representations length $n_t$ and $n_f$, remains constant and irrelevant to the input series length $L$. For example, the maximum $L$ in the experiments is 512. However, the maximum $\hat{L}$ is only 48, since we use at most 16 frequency components and 32 time-domain patches. The following analysis shows that the reduction of sequence length allows JTFT to have low complexity. The time and space complexity are not specified in the analysis since they are the same for all the main modules in JTFT. We also ignore the channel number $D$ because all the complexity analyzed is proportional to it.

In JTFT, there are three main functional modules: preprocessing, encoder, and prediction head. The preprocessing, which comprises normalization, patching, CDCT, and embedding, has a complexity of $O(\hat{L}L)=O(L)$. The Transformer encoder takes most of the computation and memory of JTFT in practice. However, its complexity is only $O(\hat{L}^2)=O(1)$, which can be a bit misleading since the coefficient is large. In contrast, the low-rank attention layers have $O(D\hat{L})=O(1)$ complexity and are much cheaper than the Transformer encoder in practical applications. The prediction head has a complexity of $O(\hat{L}T)=O(T)$, which can be approximated as $O(L)$ under normal conditions where the target window $T$ and look-back window $L$ have the same order of magnitude. Consequently, the overall complexity of JTFT is $O(L)$.

\begin{table}
	\caption{Complexity analysis of different forecasting models}
	\centering
	\label{TAB_COMPLEXITY}
	\centering
	\resizebox{\linewidth}{!}{
	\begin{tabular}{llllll}
		\toprule
		Methods     	&Time     		&Space 				&Methods     	&Time     		&Space \\
		\midrule
		Autoformer 		&$O(L\log(L))$ 	&$O(L\log(L))$  	&Crossformer	&$O(L^2)$ 		&$O(L^2)$\\
		FEDformer 		&$O(L)$ 		&$O(L)$  			&Informer		&$O(L\log(L))$ 	&$O(L\log(L))$\\
		JTFT	 		&$O(L)$ 		&$O(L)$				&LogTrans		&$O(L\log^2(L))$ 	&$O(L\log^2(L))$\\
		LSTM			&$O(L)$ 		&$O(L)$ 			&MICN			&$O(L)$ 		&$O(L)$\\
		PatchTST		&$O(L^2)$ 		&$O(L^2)$			&Pyraformer		&$O(L)$ 		&$O(L)$\\
		Reformer		&$O(L\log(L))$ 	&$O(L\log(L))$		&Transformer	&$O(L^2)$ 		&$O(L^2)$\\
		\bottomrule
	\end{tabular}
	}
\end{table}

Table \ref{TAB_COMPLEXITY} compares the time and space complexity of various time series forecasting models. JTFT is one of the prediction models with the lowest complexity ($O(L)$). Notably, Crossformer and PatchTST have high theoretical complexities ($O(L^2)$), but they effectively reduce actual computational costs by using segment projections as input to Transformer encoders instead of individual time points. 

\section{Experiments}

\textbf{Datasets:} We assess the performance of JTFT on six real-world datasets, including Exchange, Weather, Traffic, Electricity (Electricity Consumption Load), ILI (Influenza-Like Illness), and ETTm2 (Electricity Transformer Temperature-minutely). The datasets are divided into training, validation, and test sets following \citet{patchtst}, with split ratios of 0.6:0.2:0.2 for ETTm2 and 0.7:0.1:0.2 for other datasets.

\textbf{Baselines:} We use several SOTA models for time series forecasting as baselines, including PatchTST \citep{patchtst}, Crossformer \citep{crossformer}, FEDformer \citep{fedformer}, FiLM \citep{film}, DLinear \citep{dlinear}, DeepTime \citep{deeptime}, TSMixer \citep{tsmixergoogle}. Classic models such as ARIMA, basic RNN/LSTM/CNN models, and some popular Transformer-based models, including LogTrans \citep{logtrans}, Reformer \citep{reformer}, Pyraformer \citep{pyraformer}, Autoformer \citep{autoformer}, and Informer \citep{informer} are not included in the main results because they exhibit relatively inferior performance, as shown in \citep{fedformer,dlinear,patchtst}. 

\textbf{Experimental Settings:} The forecasting length for ILI is $T \in \{24, 36, 48, 60\}$, while for the other datasets, it is $T \in \{96, 192, 336, 720\}$. Baseline results for PatchTST, DLinear, FiLM, DeepTime, and TSMixer are obtained from their original papers, with the look-back window $L$ either searched for or set to a suggested value. Specific to PatchTST, we present the results of PatchTST/64, which generally performs better overall than PatchTST/42. For other methods or settings not available in the literature, $L$ is determined through grid search to establish strong baselines. The search range is $\{24, 48, 84, 96, 128\}$ for the two smaller datasets (ILI and Exchange) and $\{48, 96, 192, 336, 512, 720\}$ for the other datasets. In contrast, the JTFT configuration aligns with PatchTST/64, where $L$ is set at 128 for the smaller datasets and 512 for the larger ones. The evaluation metrics reported include Mean Squared Error (MSE) and Mean Absolute Error (MAE) for multivariate time series forecasting.

\subsection{Main results}

\linespread{1.2}
\begin{table*}[t]
	\caption{Multivariate long-term series forecasting results on 6 datasets. The best results are highlighted in \textbf{bold}, and the second best results are \underline{underlined}.}
	\centering
	\resizebox{\linewidth}{!}{
	\begin{tabular}{ccc cc cc cc cc cc cc cc ccc}
		\cline{2-19}
		&\multicolumn{2}{c}{Models}& \multicolumn{2}{c}{JTFT}& \multicolumn{2}{c}{PatchTST}&\multicolumn{2}{c}{Crossformer}& \multicolumn{2}{c}{FEDformer}& \multicolumn{2}{c}{FiLM}& \multicolumn{2}{c}{DLinear}& \multicolumn{2}{c}{DeepTime}& \multicolumn{2}{c}{TSMixer} & \\
		\cline{2-19}
		&\multicolumn{2}{c}{Metric}&MSE&MAE&MSE&MAE&MSE&MAE&MSE&MAE&MSE&MAE&MSE&MAE&MSE&MAE&MSE&MAE\\
		\cline{2-19}
		&\multirow{5}*{\rotatebox{90}{Exchange}}& 96 & \textbf{0.080} & \underline{0.199} & 0.085 & 0.202 & 0.285 & 0.410 & 0.106 & 0.245 & 0.086 & 0.204 & \underline{0.081} & 0.203 & \underline{0.081} & 0.205 & \underline{0.081} & \textbf{0.198}   \\
		&\multicolumn{1}{c}{}& 192 & \textbf{0.148} & \textbf{0.279} & 0.178 & 0.299 & 0.536 & 0.544 & 0.214 & 0.357 & 0.188 & 0.292 & 0.157 & 0.293 & \underline{0.151} & \underline{0.284} & 0.176 & 0.297   \\
		&\multicolumn{1}{c}{}& 336 & \textbf{0.260} & \textbf{0.381} & 0.329 & 0.415 & 0.804 & 0.731 & 0.413 & 0.493 & 0.356 & 0.433 & \underline{0.305} & 0.414 & 0.314 & \underline{0.412} & 0.334 & 0.416   \\
		&\multicolumn{1}{c}{}& 720 & \underline{0.667} & \underline{0.618} & 0.901 & 0.715 & 1.266 & 0.926 & 1.038 & 0.796 & 0.727 & 0.669 & \textbf{0.643} & \textbf{0.601} & 0.856 & 0.663 & 0.867 & 0.702   \\
		&\multicolumn{1}{c}{}& avg & \textbf{0.289} & \textbf{0.369} & 0.373 & 0.408 & 0.723 & 0.653 & 0.443 & 0.473 & 0.339 & 0.400 & \underline{0.296} & \underline{0.378} & 0.350 & 0.391 & 0.364 & 0.403   \\
		\cline{2-19}
		&\multirow{5}*{\rotatebox{90}{Weather}}& 96   & \textbf{0.144} & \textbf{0.186} & 0.149 & \underline{0.198} & 0.148 & 0.212 & 0.238 & 0.314 & 0.199 & 0.262 & 0.176 & 0.237 & 0.166 & 0.221 & \underline{0.145} & \underline{0.198}   \\
		&\multicolumn{1}{c}{}& 192  & \textbf{0.187} & \textbf{0.228} & 0.194 & \underline{0.241} & \underline{0.191} & 0.258 & 0.275 & 0.329 & 0.228 & 0.288 & 0.22 & 0.282 & 0.207 & 0.261 & \underline{0.191} & 0.242    \\
		&\multicolumn{1}{c}{}& 336  & \textbf{0.237} & \textbf{0.270} & 0.245 & 0.282 & 0.244 & 0.308 & 0.339 & 0.377 & 0.267 & 0.323 & 0.265 & 0.319 & 0.251 & 0.298 & \underline{0.242} & \underline{0.280}   \\
		&\multicolumn{1}{c}{}& 720  & \underline{0.308} & \textbf{0.321} & 0.314 & \underline{0.334} & 0.311 & 0.355 & 0.389 & 0.409 & 0.319 & 0.361 & 0.323 & 0.362 & \textbf{0.301} & 0.338 & 0.320 & 0.336  \\
		&\multicolumn{1}{c}{}& avg  & \textbf{0.219} & \textbf{0.251} & 0.226 & \underline{0.264} & \underline{0.224} & 0.283 & 0.310 & 0.357 & 0.253 & 0.308 & 0.246 & 0.300 & 0.231 & 0.280 & 0.225 & \underline{0.264}   \\
		\cline{2-19}
		&\multirow{5}*{\rotatebox{90}{Traffic}}& 96   & \textbf{0.351} & \textbf{0.232} & \underline{0.360} & \underline{0.249} & 0.482 & 0.268 & 0.576 & 0.359 & 0.416 & 0.294 & 0.410 & 0.282 & 0.390 & 0.275 & 0.376 & 0.264    \\
		&\multicolumn{1}{c}{} & 192  &  \textbf{0.374} & \textbf{0.243} & \underline{0.379} & \underline{0.256} & 0.495 & 0.271 & 0.610 & 0.380 & 0.408 & 0.288 & 0.423 & 0.287 & 0.402 & 0.278 & 0.397 & 0.277   \\
		&\multicolumn{1}{c}{}& 336   &  \textbf{0.385} & \textbf{0.249} & \underline{0.392} & \underline{0.264} & 0.512 & 0.280 & 0.608 & 0.375 & 0.425 & 0.298 & 0.436 & 0.296 & 0.415 & 0.288 & 0.413 & 0.290   \\
		&\multicolumn{1}{c}{}& 720   &  \textbf{0.429} & \textbf{0.275} & \underline{0.432} & \underline{0.286} & 0.561 & 0.313 & 0.621 & 0.375 & 0.52 & 0.353 & 0.466 & 0.315 & 0.449 & 0.307 & 0.444 & 0.306   \\
		&\multicolumn{1}{c}{}& avg   &  \textbf{0.385} & \textbf{0.250} & \underline{0.391} & \underline{0.264} & 0.512 & 0.283 & 0.604 & 0.372 & 0.442 & 0.308 & 0.434 & 0.295 & 0.414 & 0.287 & 0.407 & 0.284  \\
		\cline{2-19}
		&\multirow{5}*{\rotatebox{90}{Electricity}}& 96  & \underline{0.131} & \underline{0.224} & \textbf{0.129} & \textbf{0.222} & 0.217 & 0.311 & 0.186 & 0.302 & 0.154 & 0.267 & 0.140 & 0.237 & 0.137 & 0.238 & \underline{0.131} & 0.229   \\
		&\multicolumn{1}{c}{}& 192   & \textbf{0.144} & \textbf{0.237} & \underline{0.147} & \underline{0.240} & 0.263 & 0.337 & 0.197 & 0.311 & 0.164 & 0.258 & 0.153 & 0.249 & 0.152 & 0.252 & 0.151 & 0.246    \\
		&\multicolumn{1}{c}{}& 336   & \textbf{0.157} & \textbf{0.252} & 0.163 & \underline{0.259} & 0.319 & 0.370 & 0.213 & 0.328 & 0.188 & 0.283 & 0.169 & 0.267 & 0.166 & 0.268 & \underline{0.161} & 0.261   \\
		&\multicolumn{1}{c}{}& 720   & \textbf{0.182} & \textbf{0.275} & \underline{0.197} & \underline{0.290} & 0.388 & 0.412 & 0.233 & 0.344 & 0.236 & 0.332 & 0.203 & 0.301 & 0.201 & 0.302 & \underline{0.197} & 0.293 \\
		&\multicolumn{1}{c}{}& avg   & \textbf{0.154} & \textbf{0.247} & \underline{0.159} & \underline{0.253} & 0.297 & 0.357 & 0.207 & 0.321 & 0.185 & 0.285 & 0.166 & 0.264 & 0.164 & 0.265 & 0.160 & 0.257   \\
		\cline{2-19}
		&\multirow{5}*{\rotatebox{90}{ILI}}& 24    &  \textbf{1.027} & \textbf{0.604} & \underline{1.319} & \underline{0.754} & 2.918 & 1.139 & 2.624 & 1.095 & 1.970 & 0.875 & 2.215 & 1.081 & 2.425 & 1.086 & 2.415 & 1.058   \\
		&\multicolumn{1}{c}{} & 36   & \textbf{0.995} & \textbf{0.621} & \underline{1.579} & 0.870 & 3.020 & 1.123 & 2.516 & 1.021 & 1.982 & \underline{0.859} & 1.963 & 0.963 & 2.231 & 1.008 & 2.280 & 1.027  \\
		&\multicolumn{1}{c}{}& 48    & \textbf{0.980} & \textbf{0.637} & \underline{1.553} & \underline{0.815} & 3.241 & 1.192 & 2.505 & 1.041 & 1.868 & 0.896 & 2.130 & 1.024 & 2.230 & 1.016 & 2.379 & 1.056  \\
		&\multicolumn{1}{c}{}& 60    & \textbf{1.386} & \textbf{0.760} & \underline{1.470} & \underline{0.788} & 3.324 & 1.188 & 2.742 & 1.122 & 2.057 & 0.929 & 2.368 & 1.096 & 2.143 & 0.985 & 2.370 & 1.047   \\
		&\multicolumn{1}{c}{}& avg    & \textbf{1.097} & \textbf{0.655} & \underline{1.480} & \underline{0.807} & 3.126 & 1.161 & 2.597 & 1.070 & 1.969 & 0.890 & 2.169 & 1.041 & 2.257 & 1.024 & 2.361 & 1.047  \\
		\cline{2-19}
		&\multirow{4}*{\rotatebox{90}{ETTm2}} & 96    & \textbf{0.160} & \textbf{0.247} & 0.166 & 0.256 & 0.280 & 0.371 & 0.180 & 0.271 & 0.165 & 0.256 & 0.167 & 0.260 & 0.166 & 0.257 & \underline{0.163} & \underline{0.252}  \\
		&\multicolumn{1}{c}{}& 192   & \textbf{0.213} & \textbf{0.284} & 0.223 & 0.296 & 0.364 & 0.446 & 0.252 & 0.318 & 0.222 & 0.296 & 0.224 & 0.303 & 0.225 & 0.302 & \underline{0.216} & \underline{0.290}  \\
		&\multicolumn{1}{c}{}& 336   & \textbf{0.265} & \textbf{0.319} & 0.274 & 0.329 & 0.990 & 0.734 & 0.324 & 0.364 & 0.277 & 0.333 & 0.281 & 0.342 & 0.277 & 0.336 & \underline{0.268} & \underline{0.324} \\
		&\multicolumn{1}{c}{}& 720   & \textbf{0.348} & \textbf{0.373} & \underline{0.362} & \underline{0.385} & 1.892 & 1.026 & 0.410 & 0.420 & 0.371 & 0.389 & 0.397 & 0.421 & 0.383 & 0.409 & 0.420 & 0.422 \\
		&\multicolumn{1}{c}{}& avg   & \textbf{0.247} & \textbf{0.306} & \underline{0.256} & \underline{0.317} & 0.881 & 0.644 & 0.291 & 0.343 & 0.259 & 0.319 & 0.267 & 0.332 & 0.263 & 0.326 & 0.267 & 0.322  \\
		\cline{2-19}
	\end{tabular}
	}
	\label{TAB_MAIN_RES}
\end{table*}
\linespread{1}

Table \ref{TAB_MAIN_RES} presents the multivariate forecasting results, where JTFT excels over all baseline methods. Across 60 different settings involving various datasets, evaluation metrics, and diverse prediction lengths, it ranks top-1 in 54 settings and top-2 in the remaining 6 settings. This performance surpasses previous FD methods like FEDformer and FiLM, as well as the SOTA Transformer-based model, PatchTST. Worth noting is that FEDformer significantly outperforms earlier well-known Transformer models such as Autoformer, Informer, LogTrans, and Performer \citep{fedformer}, and FiLM is an improved iteration of FEDformer proposed by the same team. The channel-independent PatchTST outperforms Crossformer, designed for extracting cross-channel dependencies. The results highlight the challenge of improving the performance of SOTA channel-independent methods by incorporating cross-channel information in multivariate time series forecasting, primarily due to the relatively smaller sizes of available datasets compared to applications like computer vision (CV) and natural language processing (NLP), which carry a higher risk of overfitting. Alongside Transformer-based methods, DLinear, FiLM, DeepTime, and TSMixer are shown to be competitive.

\subsection{Ablation study}

In this subsection, we investigate the impact of the Joint Time-Frequency Domain Representation (JTFR) and Low-Rank Attention (LRA) layers. We refer to JTFT with only a TD representation and no LRA as TDR, JTFT with only an FD representation and no LRA as FDR, and the one with both TD and FD representations but no LRA as JTFR. The performance of TDR, FDR, and JTFR is assessed on three datasets: ILI, Weather, and Electricity, representing small, medium, and large datasets, respectively. The results are compared with PatchTST in Table \ref{TAB_ABLATION}. FEDformer and FiLM are also included in the baseline as previous SOTA FD methods, which utilize random and low frequencies, respectively.

Table \ref{TAB_ABLATION} demonstrates that JTFR enhances performance in comparison to both TDR and FDR across most settings. The inclusion of both TD and FD representations provides additional information to correct errors in predictions made using only one of them. Specifically, JTFR achieves performance comparable to, or in some cases, slightly better than PatchTST while reducing the length of the representation $\hat{L}$. Here, $\hat{L}$ corresponds to the patch number in PatchTST and is the sum of TD and FD patch numbers ($n_t$ and $n_f$) in JTFT. In the experiment, $\hat{L}$ is reduced from 64 to 32 in Weather and 48 in ILI and Electricity, reducing the computation and memory costs since the complexity of Transformers is proportional to $O(\hat{L}^2)$. Additionally, FDR significantly outperforms FEDformer and even surpasses FiLM in most of the settings, highlighting the effectiveness of learnable frequencies. JTFT outperforms JTFR, showcasing that LRA inclusion enhances prediction performance by leveraging cross-channel correlations.

\linespread{1.2}
\begin{table*}[t]
	\caption{Ablation study of the Joint Time-Frequency Domain Representation (JTFR) and Low-Rank Attention Layer (LRA) in JTFT. TDR denotes JTFT with only a TD representation and no LRA, FDR represents JTFT with only an FD representation and no LRA, and JTFR includes both TD and FD representations but no LRA. The best results are highlighted in \textbf{bold}, and the second-best results are \underline{underlined}.}
	\centering
	\resizebox{\linewidth}{!}{
		\begin{tabular}{ccc cc cc cc cc cc cc cc}
			\cline{2-17}
			&\multicolumn{2}{c}{Models}	&\multicolumn{2}{c}{JTFT}
			&\multicolumn{2}{c}{TDR}	&\multicolumn{2}{c}{FDR} 
			&\multicolumn{2}{c}{JTFR}	&\multicolumn{2}{c}{FEDformer}
			&\multicolumn{2}{c}{FiLM}	&\multicolumn{2}{c}{PatchTST} \\
			\cline{2-17}
			&\multicolumn{2}{c}{Metric}&MSE&MAE&MSE&MAE&MSE&MAE&MSE&MAE&MSE&MAE&MSE&MAE&MSE&MAE\\
			\cline{2-17}
			&\multirow{5}*{\rotatebox{90}{ILI}}& 24
			&\textbf{1.027} &\textbf{0.604}	&1.509 &0.718	&1.626 &0.835	&\underline{1.124} &\underline{0.668}
			& 2.624 & 1.095 					&1.970 &0.875
			&1.319 &0.754   \\
			&\multicolumn{1}{c}{} & 36   
			&\textbf{0.995} &\textbf{0.621}	&1.449 &0.715	&1.739 &0.887	&\underline{1.055} &\underline{0.653}
			& 1.153 & 0.679   					&1.982 &0.859
			&1.579 &0.870\\
			&\multicolumn{1}{c}{}& 48    
			&\textbf{0.980} &\textbf{0.637}	&1.490 &0.738	&1.864 &0.957	&\underline{1.107} &\underline{0.690}
			& 2.505 & 1.041   					&1.868 &0.896
			&1.553 &0.815\\
			&\multicolumn{1}{c}{}& 60    
			&\textbf{1.386} &\textbf{0.760}	&1.818 &0.846	&1.946 &0.993	&\underline{1.454} &0.794
			& 2.742 & 1.122   					&2.057 &0.929
			& 1.470 &\underline{0.788}\\
			&\multicolumn{1}{c}{}& avg 
			&\textbf{1.097}	&\textbf{0.656} 	&1.567 	&0.754 	&1.794 	&0.918 	&\underline{1.185} 	&\underline{0.701} 	&2.256 	&0.984 	&1.969 	&0.890 	&1.480 	&0.807 
			\\
			\cline{2-17}
			&\multirow{5}*{\rotatebox{90}{Weather}}& 96   
			&\textbf{0.144} &\textbf{0.186}	&0.158 &0.201	&\underline{0.147} &0.193	&\underline{0.147} &\underline{0.190} 
			& 0.238 & 0.314    					&0.199 &0.262
			& 0.149 & 0.198 \\
			&\multicolumn{1}{c}{}& 192  
			&\textbf{0.187} &\textbf{0.228}	&0.200 &0.240	&\underline{0.191} &0.237	&0.192 &\underline{0.234} 
			& 0.275 & 0.329    					&0.228 &0.288
			& 0.194 & 0.241  \\
			&\multicolumn{1}{c}{}& 336  
			&\textbf{0.237} &\textbf{0.270}	&0.251 &0.280	&\underline{0.245} &0.280	&\underline{0.245} &\underline{0.276} 
			& 0.339 & 0.377   					&0.267 &0.323
			& \underline{0.245} & 0.282  \\
			&\multicolumn{1}{c}{}& 720  
			&\textbf{0.308} &\textbf{0.321}	&0.322 &0.332	&0.319 &0.334	&0.316 &\underline{0.328}	
			& 0.389 & 0.409   					&0.319 &0.361
			& \underline{0.314} & 0.334\\
			&\multicolumn{1}{c}{}& avg
			&\textbf{0.219} 	&\textbf{0.251} 	&0.233 	&0.263 	&0.226 	&0.261 	&\underline{0.225} 	&\underline{0.257} 	&0.310 	&0.357 	&0.253 	&0.309 	&0.226 	&0.262 
			\\
			\cline{2-17}
			&\multirow{5}*{\rotatebox{90}{Electricity}}& 96   
			&\underline{0.131} &0.224	&0.132 &\textbf{0.221}	&0.146 &0.245	&\underline{0.131} &\underline{0.222}	 
			& 0.186 & 0.302   						&0.154 &0.267
			& \textbf{0.129} & \underline{0.222}\\
			&\multicolumn{1}{c}{}& 192   
			&\textbf{0.144} &\textbf{0.237}	&0.148 &\textbf{0.237}	&0.161 &0.259	&\underline{0.147} &\textbf{0.237}
			& 0.197 & 0.311  						&0.164 &0.258
			& \underline{0.147} & 0.240\\
			&\multicolumn{1}{c}{}& 336   
			&\textbf{0.157} &\textbf{0.252}	&0.164 &0.255	&0.176 &0.272	&\underline{0.163} &\underline{0.253}	
			& 0.213 & 0.328  						&0.188 &0.283
			& \underline{0.163} & 0.259\\
			&\multicolumn{1}{c}{}& 720   
			&\textbf{0.182} &\textbf{0.275}	&0.205 &0.288	&0.211 &0.300	&0.200 &\underline{0.286}	 
			& 0.233 & 0.344 			 			&0.236 &0.332
			& \underline{0.197} & 0.290 \\
			&\multicolumn{1}{c}{}& avg
			&\textbf{0.154} 	&\textbf{0.247} 	&0.162 	&0.250 	&0.174 	&0.269 	&0.160 	&\underline{0.250} 	&0.207 	&0.321 	&0.186 	&0.285 	&\underline{0.159} 	&0.253 
			\\
			\cline{2-17}
		\end{tabular}
	}
	\label{TAB_ABLATION}
\end{table*}
\linespread{1}

\subsection{Actual time efficiency and memory usage}

In some cases, methods with low theoretical complexity may still incur significant costs in practice due to large coefficients in the expressions. In this subsection, we compare the actual time and memory costs of JTFT with other Transformer-based methods (e.g., PatchTST, FEDformer, Crossformer, Autoformer, Informer) and DLinear.

The experiments were conducted on the Weather dataset using an NVIDIA RTX 3090 GPU. All previous methods used the settings detailed in their original papers. To ensure a fair comparison of computational efficiency and memory usage, the batch size was uniformly set to 128. However, for Autoformer and Informer, a batch size of 64 was used for look-back windows of 96, 192, and 336, respectively, due to GPU memory limitations. Tests were not performed on the two methods with larger look-back windows, as doing so would have necessitated a further reduction in batch sizes, potentially resulting in an unfair comparison of speed. Additionally, the compilation function in PyTorch 2 was disabled.

\begin{figure}
	\centering
	\includegraphics[width=\linewidth]{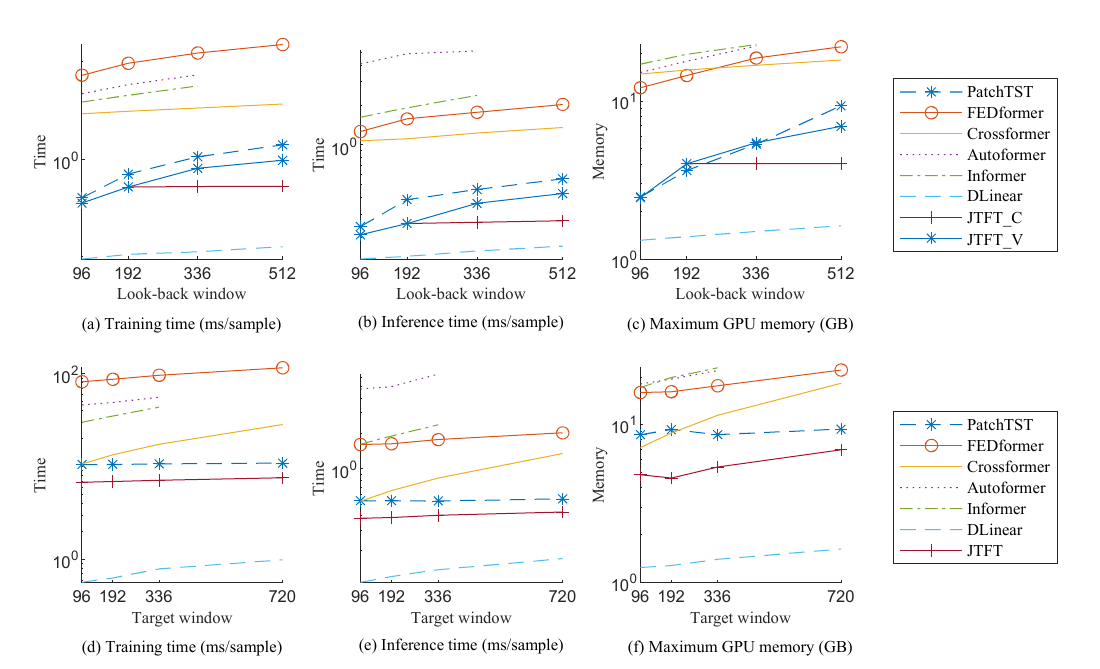}
	\caption{Comparison of actual runtime efficiency and memory usage. Figures (a, b, c) depict the results for various look-back windows while the target window is 720. Figures (d, e, f) explore the impact of target window variations when the look-back window is 512. In JTFT\_C, the TD and FD representation lengths remain constant, while in JTFT\_V, the representation lengths increase roughly linearly with the look-back window. In Figures (d, e, f), JTFT uses the largest representation length of JTFT\_V. In general, JTFT is more efficient in both computation and memory than the other Transformer-based methods in the comparison.}
	\label{FIG_COSTS}
\end{figure}

The results are displayed in Figure \ref{FIG_COSTS}. Among the Transformer-based methods, JTFT is the fastest and requires the least memory. While DLinear is faster than JTFT, its accuracy is inferior. These results highlight that JTFT is an efficient approach, especially when considering both speed and accuracy. Further details and explanations are provided below.

Firstly, the computation and memory costs for different look-back windows $L$ are compared in Figures \ref{FIG_COSTS} (a, b, c), where the target window size consistently set at 720. According to the complexity analysis, although the time and space complexity of JTFT is $O(L)$, the real time and memory cost are mainly decided by the input length of the Transformer encoder, denoted by $\hat{L}=n_t+n_f$. In the expression, $n_t$ and $n_f$ are the length of TD and FD representation, which are assigned by users and decoupled from $L$. Typically, $\hat{L}$ is less than the number of patches (refer to Appendix A), since the FD and TD representation are linearly dependent otherwise.

Two versions of JTFT with different strategies to set $n_t$ and $n_f$ are considered. In JTFT\_C uses constant $(n_t,n_f)=(8,8)$. It is not applied for $L=96$ due to the number of patches being less than $\hat{L}$. In JTFT\_V, $(n_t,n_f)$ increases along with $L$. It is set as (4, 4), (8, 8), (12, 12), and (16, 16) for $L$ of 96, 192, 336, and 512. 

Figures \ref{FIG_COSTS} (a, b, c) demonstrate that both JTFT\_C and JTFT\_V outperform the other Transformer-based methods in most settings. The computational time and GPU memory usage of JTFT\_C remain almost constant with respect to $L$ because the costs of encoders and heads, which account for most of the expenses in JTFT, do not increase with $L$ when $\hat{L}$ is kept constant. In contrast, for JTFT\_V, the computation and memory costs increase as $\hat{L}$ grows with $L$. JTFT\_C and JTFT\_V require slightly more memory compared to PatchTST when $L=192$ because they include additional LRA layers.

Next, we compare the computational and memory costs for various target windows ($T$) in Figure \ref{FIG_COSTS} (d, e, f). The look-back window size is consistently set to 512 in these figures. Similar to the previous results, JTFT remains faster and requires less memory compared to the other Transformer-based methods. The costs increase at a small and roughly constant rate with respect to $T$. This behavior is because, when both $L$ and $\hat{L}$ are fixed, only the cost of the prediction head, which is not resource-intensive, increases along with $T$.

\section{Conclusions and future works}

This paper introduces JTFT, a joint time-frequency domain Transformer for multivariate time series forecasting. JTFT effectively captures multi-scale structures using a small number of learnable frequencies, while also leveraging the latest time-domain data to enhance local relation learning and mitigate the adverse effects of non-stationarity. Additionally, it utilizes a low-rank attention layer to extract cross-channel correlations while alleviating the entanglement with the modeling of temporal dependencies. JTFT has linear complexity in both time and space. Extensive experiments on 6 real-world datasets demonstrate that our method achieves state-of-the-art performance in long-term forecasting.

In addition to its application in the Transformer architecture, the joint time-frequency domain representation could be utilized in other types of neural network models, including those based on CNNs and MLPs. Moreover, this representation holds the potential for adoption in various domains, such as NLP, where input sequences tend to be lengthy.
	
\section{Acknowledgments}

This work was supported in part by the National Natural Science Foundation of China (Grant No. T2125006, U2242210, 42174057, 61972231, 62102114, 62202119), the Jiangsu Innovation Capacity
Building Program (Grant No. BM2022028), the Science and Technology Project of Qinghai Province (Grant No. 2023-QY-208), and the Key Research Project of Zhejiang Lab (Grant No. 2021PB0AC01).

\appendix

\section{A brief introduction about patching in time series forecasting}

As analyzed in Section 3, the patching technique and channel-independent setting introduced in PatchTST have been found to be effective in extracting temporal dependencies, and are thus applied in JTFT. We further propose the time-frequency-independence setting to exploit the cross-channel dependencies with low redundancy after the temporal modeling. Although these approaches are not inherently complex, they may not be intuitive for readers who are unfamiliar with PatchTST. Therefore, some brief explanations are presented below.

Patching involves dividing the time series into either overlapped or non-overlapped continuous segments, which serve as the fundamental input units for subsequent modeling steps. This technique retains local semantic information within the embedding, thereby enhancing the capture of comprehensive semantic information that may not be available at the point-level. Moreover, it reduces the length of inputs for Transformer encoders, resulting in significant computational and memory savings.

During the patching process, each channel of the input multivariate time series is treated as a separate univariate time series. The input univariate time series are then divided into patches, with parameters defined as follows:  the patch length is denoted as $l_p$, and the stride as $l_s$. It is assumed that the input length $L$ and patch length $l_p$ are divisible by $l_s$, and the time series is padded by repeating the last element $l_s$ times. In this setup, a total of $\bar{L}=(L-l_p)/l_s +2$ patches are generated by extracting continuous segments within the input sequence. The input is rearranged as 
\begin{equation}
	\text{Patching}\left(\mathbf{x}_{1:L}\right)  =\mathbf{Z}_{pch}
\end{equation}
where the input series $\mathbf{x}_{1:L} \in \mathbb{R}^{D \times L}$, patched series $\mathbf{Z}_{pch} \in \mathbb{R}^{D \times \bar{L} \times l_p}$, and
\begin{equation}
	\left(\mathbf{Z}_{pch}\right)^i_j=  \mathbf{x}_{(j-1)l_s+1:(j-1)l_s+l_p}^i
\end{equation}
for channel index $i \in \{1,\cdots, D\}$ and patch index $j \in \{1,\cdots, \hat{L}\}$.

Patching reduces the input sequence length of the Transformer encoder from $L$ to approximately $L/l_s$, that results in significant saving in computation and memory. This reduction also allows for more efficient processing of long sequences.

The channel-independent (CI) setting treats each channel of a multivariate series as an independent univariate series. The network structures under the CI setting are shared for all channels, so that much more data samples are available in training. Consequently, CI networks are more stable to noise and less prone to overfitting. In implementation, the patched data is permuted and the channel dimension is merged into the batch-size dimension, that enables the data to be processed by a vanilla Transformer encoder directly. 

The overall structure of PatchTST is under the CI setting, which brings an  enormous advantage compared with the channel-mixed models and achieves SOTA performance. However, the trade-off of the CI setting is that it is unable to capture cross-channel dependencies. 

In order to utilize the cross-channel relations, we apply a time-frequency-independence (TFI) setting in our low-rank attention (LRA) layers. It is proposed based on the observation that incorporating CI and non-CI network structures typically leads to degeneration of performance compared with the strong baseline of PatchTST. TFI shares the network structures along the time-frequency dimension instead of the channel dimension in CI. Furthermore, being more compact, it also shares the increments resulting from cross-channel interactions across the time-frequency dimension. This approach significantly reduces the updating space in LRA, thereby mitigating the entanglement and redundancy associated with modeling temporal and channel-wise relations simultaneously.

JTFT successfully combines the CI and CFI settings. The Transformer encoder and its embedding are channel-independent, resulting in a CI representation of the input series. The representation is fed to the TFI LRA layers to incorporate cross-channel information. Finally, the prediction head maps the representation to prediction. The head is also shared along the channel dimension, but the channel-wise information has been integrated into its inputs. Empirical results show that the mixed CI\_TFI design is effective in most datasets.

\section{Datasets details}

The details of the datasets are introduced as follows:
(1) ETTm2 \citep{informer} contains the electricity transformer temperature and power loads collected from 2 counties in China. It is a 15-minute-level dataset spanning 2 years.
(2) Electricity \footnote{\url{https://archive.ics.uci.edu/ml/datasets/ElectricityLoadDiagrams20112014}} comprises the hourly electricity consumption of 321 customers, also spanning 2 years.
(3) Exchange \citep{Lai2018} logs the daily exchange rates of eight different countries from 1990 to 2016.
(4) ILI \footnote{\url{https://gis.cdc.gov/grasp/fluview/fluportaldashboard.html}} includes the weekly data of recorded influenza-like illness (ILI) patients from the Centers for Disease Control and Prevention of the United States, covering the period from 2002 to 2021. This dataset provides information on the ratio of patients seen with ILI to the total number of patients.
(5) Traffic \footnote{\url{http://pems.dot.ca.gov/}} consists of hourly data obtained from the California Department of Transportation, providing information on road occupancy rates measured by various sensors installed on freeways in the San Francisco Bay area.
(6) Weather \footnote{\url{https://www.bgc-jena.mpg.de/wetter/}} is captured at 10-minute intervals throughout the year 2020, featuring 21 meteorological indicators, such as air temperature, humidity, and wind speed. The statistics of the datasets are summarized in Table \ref{TAB_DATASET}. All the datasets can be accessed from \href{https://github.com/thuml/Autoformer}{Autoformer}.

\begin{table}
	\caption{Statistics of the datasets for benchmark}
	\label{TAB_DATASET}
	\centering
	\begin{tabular}{lllllll}
		\toprule
		Dataset   &Ettm2      &Electricity    &Exchange   &ILI    &Traffic    &Weather \\
		\midrule
		Length 	  &69980      &26304          &7588       &966    &17544      &52696\\
		Channels  &7          &321            &8          &7      &862        &21\\
		Frequency &15 min     &1 hour         &1 day      &7 day  &1 hour     &10 min\\  
		\bottomrule
	\end{tabular}
\end{table}

\section{Experimental details}

By default, JTFT consists of 3 Transformer encoder layers and 1 low-rank attention layer (LRA). However, the number of LRA is set to 0 for the Traffic dataset as it negatively impacts the performance. The dimension of the latent space, denoted as $d_m$, in the Transformer encoder increases with the dataset size. Specifically, it is set to 8 for ILI and Exchange, 16 for ETTm2, 128 for Weather, Electricity and Traffic. The $d_m$ of the LRA is half of those in the Transformer encoder. For ILI and Exchange, the patch length and stride are set to $(4,2)$, while for the other datasets, they are $(16,8)$. The length of TD and FD representations $(n_t,n_f)$ are searched in $(16,16)$ and $(16,32)$.

Our method is implemented in PyTorch and trained on a workstation equipped with 4 NVIDIA RTX 3090 GPUs, each with 24GB memory. All 4 GPUs are utilized for training on the Electricity and Traffic datasets, while only 1 GPU is used for the remaining datasets.

\section{Multiple random runs}

The results of JTFT reported in the main text are obtained using a fixed random seed of 1. To assess the robustness of the method, we conduct 5 random runs on the ILI dataset, and 3 runs on the Weather and Traffic datasets. These datasets represent small, medium, and large datasets, respectively. PatchTST, which performed the best among the previous experiments except for JTFT, is included as a baseline, along with Dlinear.

\linespread{1.2}
\begin{table*}[t]
	\caption{Multivariate long-term series forecasting results on 3 datasets showing both Mean and STD. The best results are highlighted in \textbf{bold}, and the second best results are \underline{underlined}.}
	\centering
	\resizebox{\linewidth}{!}{
		\begin{tabular}{ccc cc cc cc}
			\cline{2-9}
			&\multicolumn{2}{c}{Dataset}  & \multicolumn{2}{c}{JTFT}   & \multicolumn{2}{c}{PatchTST} & \multicolumn{2}{c}{DLinear}\\
			\cline{2-9}
			&\multicolumn{2}{c}{Metric}&MSE&MAE&MSE&MAE\\
			\cline{2-9}
			&\multirow{4}*{\rotatebox{90}{ILI}}& 24
			&$\mathbf{0.9940\pm0.0461}$  &$\mathbf{0.6053\pm0.0064}$	
			&\underline{$1.3989\pm0.0844$} &\underline{$0.7670\pm0.0233$}
			&$1.9569\pm0.0232$ &$0.9788\pm0.0070$\\
			&\multicolumn{1}{c}{} & 36   
			&$\mathbf{1.0380\pm0.0744}$  &$\mathbf{0.6414\pm0.0191}$ 	        
			&\underline{$1.2534\pm0.0778$} &\underline{$0.7381\pm0.0240$}
			&$2.0823\pm0.0067$ &$0.9980\pm0.0024$			\\
			&\multicolumn{1}{c}{}& 48    
			&$\mathbf{1.0639\pm0.0597}$  &$\mathbf{0.6639\pm0.0174}$            
			&\underline{$1.6462\pm0.1520$} &\underline{$0.8318\pm0.0513$}		
			&$2.1333\pm0.0063$ &$1.0251\pm0.0022$\\
			&\multicolumn{1}{c}{}& 60    
			&$\mathbf{1.3866\pm0.0568}$  &$\mathbf{0.7593\pm0.0152}$ 	        
			&\underline{$1.4527\pm0.0547$} &\underline{$0.8008\pm0.0105$}
			&$2.3175\pm0.0178$ &$1.0854\pm0.0090$ \\
			\cline{2-9}
			&\multirow{4}*{\rotatebox{90}{Weather}}& 96
			&$\mathbf{0.1440\pm0.0003}$  &$\mathbf{0.1875\pm0.0016}$	    
			&\underline{$0.1483\pm0.0004$} &\underline{$0.1978\pm0.0002$}		   
			&$0.1690\pm0.0004$ &$0.2300\pm0.0015$\\
			&\multicolumn{1}{c}{} & 192   
			&$\mathbf{0.1876\pm0.0008}$  &$\mathbf{0.2300\pm0.0016}$
			&\underline{$0.1943\pm0.0013$} &\underline{$0.2419\pm0.0013$}			
			&$0.2133\pm0.0012$ &$0.2724\pm0.0033$\\
			&\multicolumn{1}{c}{}& 336   
			&$\mathbf{0.2385\pm0.0013}$  &$\mathbf{0.2710\pm0.0013}$      
			&\underline{$0.2461\pm0.0010$} &\underline{$0.2828\pm0.0007$}		
			&$0.2570\pm0.0016$ &$0.3078\pm0.0030$ \\
			&\multicolumn{1}{c}{}& 720   
			&$\mathbf{0.3087\pm0.0008}$  &$\mathbf{0.3233\pm0.0019}$ 	    
			&\underline{$0.3126\pm0.0011$} &\underline{$0.3329\pm0.0011$}    
			&$0.3161\pm0.0016$ &$0.3559\pm0.0026$\\
			\cline{2-9}
			&\multirow{4}*{\rotatebox{90}{Traffic}}& 96
			&$\mathbf{0.3525\pm0.0021}$  &$\mathbf{0.2339\pm0.0016}$    
			&\underline{$0.3602\pm0.0006$} &\underline{$0.2487\pm0.0003$}
			&$0.4101\pm0.0001$ &$0.2819\pm0.0002$\\
			&\multicolumn{1}{c}{} & 192   
			&$\mathbf{0.3732\pm0.0005}$  &$\mathbf{0.2425\pm0.0002}$
			&\underline{$0.3788\pm0.0004$} &\underline{$0.2560\pm0.0002$}
			&$0.4227\pm0.0003$ &$0.2873\pm0.0002$\\
			&\multicolumn{1}{c}{}& 336   
			&$\mathbf{0.3854\pm0.0013}$  &$\mathbf{0.2500\pm0.0008}$ 
			&\underline{$0.3917\pm0.0011$} &\underline{$0.2639\pm0.0007$}
			&$0.4357\pm0.0003$ &$0.2956\pm0.0004$\\
			&\multicolumn{1}{c}{}& 720    
			&$\mathbf{0.4292\pm0.0006}$  &$\mathbf{0.2753\pm0.0005}$  	 
			&\underline{$0.4322\pm0.0009$} &\underline{$0.2863\pm0.0003$}
			&$0.4658\pm0.0001$ &$0.3148\pm0.0001$\\
			\cline{2-9}
		\end{tabular}
	}
	\label{TAB_MULTI_RUN}
\end{table*}
\linespread{1}

The mean and standard deviation of the MSE and MAE are reported in Table \ref{TAB_MULTI_RUN}. The results demonstrate that the variances of JTFT are low, particularly for the Weather and Traffic datasets. Slightly higher variances are observed for ILI, which can be attributed to the smaller size of the dataset. The comparison also reveals that JTFT significantly outperforms PatchTST and Dlinear in the majority of experimental settings, as the improvements in the mean metrics are much larger than the standard deviations. While DLinear exhibits inferior performance compared to JTFT and PatchTST, it displays remarkably low variance on ILI and Traffic. This characteristic may stem from the model's lower number of parameters compared to other methods.

\bibliographystyle{model5-names} 
\bibliography{myref}
	
\end{document}